\documentclass[10pt,twocolumn,letterpaper]{article}

\usepackage{iccv}              
\usepackage{multirow}

\definecolor{iccvblue}{rgb}{0.21,0.49,0.74}
\usepackage[pagebackref,breaklinks,colorlinks,allcolors=iccvblue]{hyperref}

\def\paperID{11} 
\def\confName{ICCVW}
\def\confYear{2025}

\title{Is It Certainly a Deepfake? \\ Reliability Analysis in Detection \& Generation Ecosystem}

\author{Neslihan Kose\\
Intel Labs
\and
Anthony Rhodes\\
Intel Labs
\and
Umur Aybars \c{C}ift\c{c}i\\
Binghamton University
\and
\.{I}lke Demir\\
Cauth AI
\and
{\tt\small \{neslihan.kose.cihangir, anthony.rhodes\}@intel.com, uciftci@binghamton.edu, ilke@cauth.ai}
}

\usepackage{tikz}
\usepackage{textcomp}
\usepackage{hyperref}
\usepackage{lipsum}

\newcommand\copyrighttext{%
  \footnotesize \textcopyright 2025 IEEE. Personal use of this material is permitted. Permission from IEEE must be obtained for all other uses, in any current or future media, including reprinting/republishing this material for advertising or promotional purposes, creating new collective works, for resale or redistribution to servers or lists, or reuse of any copyrighted component of this work in other works.}
\newcommand\copyrightnotice{%
\begin{tikzpicture}[remember picture,overlay]
\node[anchor=south,yshift=10pt] at (current page.south) {\fbox{\parbox{\dimexpr\textwidth-\fboxsep-\fboxrule\relax}{\copyrighttext}}};
\end{tikzpicture}%
}

\begin{document}
\maketitle
\copyrightnotice
\begin{abstract}

As generative models are advancing in quality and quantity for creating synthetic content, deepfakes begin to cause online mistrust. Deepfake detectors are proposed to counter this effect, however, misuse of detectors claiming fake content as real or vice versa further fuels this misinformation problem. We present the first comprehensive uncertainty analysis of deepfake detectors, systematically investigating how generative artifacts influence prediction confidence. As reflected in detectors' responses, deepfake generators also contribute to this uncertainty as their generative residues vary, so we cross the uncertainty analysis of deepfake detectors and generators. Based on our observations, the uncertainty manifold holds enough consistent information to leverage uncertainty for deepfake source detection. Our approach leverages Bayesian Neural Networks and Monte Carlo dropout to quantify both aleatoric and epistemic uncertainties across diverse detector architectures. We evaluate uncertainty on two datasets with nine generators, with four blind and two biological detectors, compare different uncertainty methods, explore region- and pixel-based uncertainty, and conduct ablation studies. We conduct and analyze binary real/fake, multi-class real/fake, source detection, and leave-one-out experiments between the generator/detector combinations to share their generalization capability, model calibration, uncertainty, and robustness against adversarial attacks. We further introduce uncertainty maps that localize prediction confidence at the pixel level, revealing distinct patterns correlated with generator-specific artifacts. Our analysis provides critical insights for deploying reliable deepfake detection systems and establishes uncertainty quantification as a fundamental requirement for trustworthy synthetic media detection.

\end{abstract}    

\section{Introduction}\label{sec:intro}

Recently, synthetic content has become a part of our daily lives with the proliferation of generative models. Specifically, human faces have always been the focus of computer vision algorithms with a growing interest, pursuing the same paradigm with generative models since the introduction of Generative Adversarial Networks~\citep{gan} (GAN) in 2014. In the intersection was born deepfakes: images, audio clips, or videos, where the actor or the action of the actor in the content is fabricated using deep generative models. 

Although synthetic content creation brought up many positive use cases, deepfakes are usually exploited in politics, entertainment, and security~\citep{DeepFakeAds2, DeepFakePorn}; causing the need for a line of defense~\citep{ucla}. Deepfake detectors are proposed to satisfy this need, however, their generalization and robustness vary depending on the intermediate signals they use, their model architecture, and the datasets they are trained on. Contemporary deepfake detection research has predominantly pursued accuracy maximization through increasingly sophisticated neural architectures and training methodologies. However, this accuracy-centric evaluation paradigm fundamentally overlooks a critical dimension: \textbf{prediction uncertainty}. As detection systems transition from controlled laboratory environments to real-world deployment scenarios, understanding when and why detectors exhibit uncertainty becomes as crucial as their peak performance capabilities. As opposed to evaluating them based on traditional accuracy and AUC metrics on many datasets, we analyze their core performance and understanding of data, by uncertainty analysis, which has not garnered much attention in the research community. Understanding model response of these deepfake detectors help compare their generalization in-the-wild, overfitting to the artifacts, performance beyond the training distribution, robustness against adversarial attacks, and effectiveness in source detection. This analysis is the key to focus on robust and reliable deepfake detectors to establish a trusted online future.

Current detection methodologies broadly fall into two paradigmatic approaches, using the existence or non-existence of priors to classify content as real or fake. These priors are either irreproducible authentic signals (e.g., corneal reflections~\citep{retina}, blood flow~\citep{FakeCatcher}, phoeneme-viseme mismatches~\citep{agarwal2020detecting}) in real data or small generative artifacts in fake data~\citep{mesonet,xception,efficientnet}, often achieving impressive performance on in-distribution benchmarks while struggling with cross-domain generalization. As per generative models, face generation~\citep{stylegan,Diffusion}, face swapping~\citep{FaceSwapq,li2019faceshifter}, and face reenactment~\citep{Prajwa2020-ACMmm,f2f} methods manipulate using different techniques operating on different facial regions. As a result, different generative models leave different residual traces behind~\citep{wang2020cnngenerated}. Those traces are directly correlated to the detector response, tying deepfake detectors and generators. This connection resumes and is observable in the uncertainty estimation of detectors, aggregated by model-specific uncertainty contributors such as architecture, signal manifold, and training data. Despite their widespread adoption across industry and academic applications, the uncertainty characteristics of these complementary approaches remain poorly understood. This knowledge gap proves particularly concerning given the high-stakes nature of deepfake detection deployments, limiting our understanding of when these systems can be trusted in production environments, where false positive classifications can irreparably damage reputations while false negatives enable malicious manipulation campaigns.

In this paper, we analyze uncertainty of various deepfake detectors in the presence of data generated by various deepfake generators. We use this analysis comprehensively to compare the robustness and reliability of detectors, to explain the detector response towards different generative sources. As the uncertainty can stem from multiple sources, multiple uncertainty measures are needed for an in-depth analysis.  
Our contributions include,
\begin{itemize} 
\item comprehensive uncertainty quantification across diverse detector and generator architectures
\item generator-specific uncertainty analysis revealing how different synthesis paradigms influence detector confidence patterns and calibration quality
\item image-, region-, and pixel-wise uncertainty comparisons of authenticity- and fakery-based deepfake detectors to provide an in-depth understanding of detection performance,
\item source detection capabilities showing how uncertainty patterns encode generator-specific signatures enabling forensic analysis beyond binary classification tasks
\item novel uncertainty visualization techniques through pixel-level confidence maps that complement existing explainability approaches and provide interpretable insights
\end{itemize}
We conduct our experiments with two uncertainty methods, on two datasets, nine generators, and six detectors (including four blind and two biological detectors). We relate generator properties to detector predictions through predictive and model uncertainty using Bayesian Neural Networks (BNNs) and through model uncertainty based on model variance using Monte-Carlo (MC) dropout approach. We compare and contrast uncertainties on both traditional deepfake detection and the more elaborate deepfake source detection tasks. 
We measure robustness in terms of parameter sensitivity and adversarial attacks. Finally, we formulate different uncertainty maps to intersect our uncertainty analysis with explainability methods to form the big picture of detector-generator relations in deepfakes.

\section{Related Work}
\label{sec:related}

\subsection{Deepfake Generation}
Deepfakes have been increasing in quality and quantity~\citep{acmsurvey}, mainly (1) creating entirely artificial faces from learned latent distributions~\citep{choi2018stargan, stylegan, mixsyn}, (2) replacing facial identity while attempting to preserve original conditions~\citep{FaceSwap-Gan, DeepFakes, li2019faceshifter}, or (3) modifying facial expressions and mouth movements while maintaining identity consistency~\citep{Prajwa2020-ACMmm, 10.1145/2816795.2818056,neuraltex}. Historically, autoregressive models~\citep{van2016conditional} (AR), Variational Autoencoders~\citep{vae} (VAE), Generative Adversarial Networks~\citep{gan} (GAN), or diffusion models~\citep{Diffusion} are used to create such manipulated content; all of which leave behind different generative residues based on the architecture, the noise, and the operations~\citep{wang2020cnngenerated, ganfingerprint}. 

\subsection{Deepfake Detection}
The arms race between generation and detection intensifies as it becomes impossible to distinguish deepfakes from real faces~\citep{tolosana2020deepfakesSurvey}. Deepfake detectors first focused on \textit{artifacts of fakery}, learning directly from data with ``blind'' detectors~\citep{mesonet,8124497,8014963,Li2020-Face-X-Ray-CVPR,shallownet,8014963,8553251,8639163,8682602,barni17,Guarnera_2020_CVPR_Workshops,Amerini2019-ICCVW}. Although they provide high accuracy on small datasets; they tend to overfit, they are easily manipulated by adversarial samples, and their generalization is limited across different domains, image transformations, and compression levels \citep{sophie,Carlini2017-SSP}. 

Another branch of deepfake detection explores authenticity signals, mostly hidden in biometric data. These detectors explore low to high level signals such as blinks~\citep{blink}, blood flow~\citep{FakeCatcher}, head-pose~\citep{headpose}, emotions~\citep{emotions}, gaze~\citep{etra}, and breathing~\citep{8553270}. These signals tend to be much inconsistent in fake videos, so the preservation of spatial, temporal, and spectral features in real videos provide an advantage for generalization over blind detectors. However, some of these inconsistencies are easily ``fixed'' in newer models~\citep{nvidiagaze}. 

The third and newest branch of deepfake detection aims to trace back the source generative model behind a given synthetic sample~\citep{Yu_2019_ICCV,8695364,ijcb,Ding2021DoesAG,motion}, following the hidden generative residue of the deep models. Some approaches even try to infer model parameters from these artifacts~\citep{asnani2021reverse-FB-MSU}.

\subsection{Uncertainty Estimation}
\label{sec:uncertainty}

Uncertainty estimation in machine learning involves quantifying the quality of predictions with respect to the confidence or to the model parameters. There are various approaches for uncertainty estimation including Bayesian~\citep{welling2011bayesian,blundell2015weight,gal2016dropout,pmlr2020BNN} and non-Bayesian~\citep{lakshminarayanan2017simple,liu2020simple,pmlr2020} methods. This important step towards evaluating prediction reliability can be designed with (1) probabilistic models to cover full probability distributions over predictions (e.g., using Bayesian Neural Networks~\citep{welling2011bayesian} (BNN)), (2) bootstrap methods to evaluate variability on controlled subsets of data or controlled subsets of the model weights (e.g., Monte-Carlo Dropout~\citep{gal2016dropout}), or (3) ensemble methods to combine multiple model predictions (e.g., Deep Ensembles~\citep{lakshminarayanan2017simple}). Tangentially, uncertainty calibration also gains attention to tune these techniques for capturing the prediction distributions as close to the sample distributions. Information theoretic approaches to use entropy and mutual information for estimating uncertainty by information gain (e.g., ~\citep{krishnan2020MOPED}) or calibration methods to align prediction probabilities to sample frequencies (e.g., ~\citep{krishnan2020AvUC, kose2022EaUC}) are widely used for this purpose.

\section{Methodology and Experimental Design} 

\subsection{Detector Architecture Selection}

Our comprehensive analysis encompasses six detector architectures representing major paradigmatic families and spanning the accuracy-efficiency trade-off space. We select these as representatives from their family of detectors to keep the number of detectors tractable (i.e., Inception~\citep{inception} is in the family of Xception~\citep{xception}, ShuffleNet~\citep{shufflenet} is in the family of MobileNet~\citep{mobilenetv2}, etc.).
\begin{itemize}[leftmargin=*]
\item ResNet18~\citep{resnet}: a generic lightweight blind detector
\item Xception~\citep{xception}: most used generic blind detector~\citep{ffbenchmark}
\item EfficientNet~\citep{efficientnet}: one of the top scoring detectors~\citep{ffbenchmark}
\item MobileNet~\citep{mobilenetv2}: a compact blind detector 
\item FakeCatcher~\citep{FakeCatcher}: an industry-adopted bio-detector~\citep{nsafbi}
\item Motion-based detector~\citep{motion}: a new bio-detector~\cite{motion}
\end{itemize}

Deepfake detection studies only fake and real classes, where fake class equals to one source subset if it is a per-generator experiment, else covers samples of all generators. Source detection studies number of generators plus one class (for real), which is formulated as classification.

\subsection{Generator Evaluation Landscape}
Although there are several deepfake datasets in the literature, there exists only two multi-source datasets with known generators, namely FaceForensics++~\citep{FF++} (FF) and FakeAVCeleb~\citep{khalid2021fakeavceleb} (FAVC). FF contains 1000 real and 5000 deepfake videos, each 1000 created by FaceSwap~\citep{FaceSwapq}, Face2Face~\citep{f2f}, Deepfakes~\citep{DeepFakes}, Neural Textures~\citep{neuraltex}, and FaceShifter~\citep{li2019faceshifter}, presenting a representative dataset covering various aforementioned face manipulation methods. FAVC contains unbalanced number of real and fake videos created by FaceSwapGAN~\citep{FaceSwap-Gan}, FSGAN~\citep{9010341}, and Wav-to-Lip~\citep{Prajwa2020-ACMmm}. As real class has the lowest number of videos (500), we balance our setup by randomly selecting 500 videos from each class. We utilize FF as our main dataset and use FAVC for generalization, using 70/30 train/test splits for all detectors. Lastly, for the adversarial robustness experiment, we use a simple adversarial generator as outlined in~\cite{sophie} on all subsets of FF where the black-box attack model is selected as the ResNet18 detector.

\subsection{Uncertainty Estimation}

For our analysis, we employ Bayesian Neural Networks \citep{welling2011bayesian} to extend deterministic deep neural network architectures to corresponding Bayesian variants in order to perform stochastic variational inference. This inference captures certainty measures that help us better understand the quality of predictions. Our implementation employs mean-field Gaussian variational families (Eq.~\ref{eq:q}).
\begin{equation}
\label{eq:q}
    q(\omega)=\mathcal{N}(\omega;\mu, diag(\sigma^2))
\end{equation}
where training optimizes the evidence lower bound, applying KL (Kullback-Leibler divergence) loss in addition to the cross entropy loss (Eq.~\ref{eq:l}), 
scaling of which can be controlled by $\beta$ ($kl_{factor}$) parameter as shown in our ablations. 
\begin{equation}
\label{eq:l}
    \mathcal{L}=\mathbb{E}_{q(\omega)}[\log p(y|x,\omega)] \mathcal{-} \beta D_{KL}[q(\omega)||p(\omega)]
\end{equation}

We quantify predictive uncertainty (predictive entropy) capturing both aleatoric and epistemic components (Eq.~\ref{eq:PU}), and model uncertainty (mutual information) computing the difference between the entropy of the mean of the predictive distribution and the mean of the entropy (Eq.~\ref{eq:MU})
  .

\begin{equation}
\label{eq:PU}
   H(y|x, D) := - \sum_{i=0}^{K-1} (p_{i\mu} \cdot log(p_{i\mu}))\\
\end{equation}

\begin{equation}
\label{eq:MU}
   I(y,\omega|x, D) := H(y|x, D) - E_{p(\omega|D)}[H(y|x, D)]
\end{equation}

where $p_{i\mu}$ is the predictive mean probability of $i^{th}$ class from $n$ MC samples and $K$ is the number of output classes. BNN conversion of all models is achieved using Bayesian-torch repo~\citep{krishnan2022bayesiantorch}. 
In order to help training convergence of models, we use MOPED method \citep{krishnan2020MOPED}, which enables initializing variational parameters from a pretrained deterministic model. During inference, multiple stochastic forward passes are performed over the network via sampling from posterior distribution of the weights (with $n$ MC samples). 

As a computationally efficient alternative, we employ MC Dropout~\citep{gal2016dropout} during inference, treating dropout masks as approximate posterior samples. Model uncertainty is quantified through prediction variance across multiple stochastic forward passes. 
Performance of both methods depend on multiple parameters, set optimally by our ablation studies. In MC dropout experiments, we report model uncertainty as the mean of the variance of sampling outputs. Finally, model calibration analyses are conducted using retention plots for deepfake detection tasks. 


\subsection{Pixel-wise Uncertainty}\label{sec:pixel}

One of the most prominent techniques in Explainable AI has been saliency maps~\citep{gradcam}, tracing the gradients back to input pixels to understand which pixels contribute more to the model's decision. Traditional saliency maps identify discriminative pixels but provide no information about \textit{how certain} this contribution is. We propose uncertainty maps to visualize this information to relate the model uncertainty back to generative artifacts on images. This duality can be thought analogous to having density plots in addition to retention plots for observing the model uncertainty with respect to its accuracy. We propose two types of maps: (1) conventional saliency maps derived from Bayesian variants of regular detectors, and (2) uncertainty maps tracing the uncertainty back to pixels of original images.

\subsubsection{Bayesian Saliency Maps}
Saliency is computed in the traditional way by calculating a weighted average of penultimate layer activation maps, however using the BNN-converted versions of the aforementioned detectors.
\begin{equation}
\begin{aligned}
    \alpha_k&=\frac{1}{n}\sum_n{y_{\max}}\left( \frac{1}{Z}\sum_{i,j}\frac{\partial y_{\max}}{\partial A_{ij}^k}\right),  \
    S=ReLU(\sum_k\alpha_k A^k)
\end{aligned}
\end{equation}
 The $\alpha_k$ activation weights are calculated as the pooled gradient magnitude of the $k^{th}$ activation map $A^k$, scaled by the predictive confidence $y_{\max}$ of the model, and averaged over the $n$ MC samples provided to the model, and computing the final saliency map $S$ by a linear combination of the $A^k$ activations with respect to $\alpha_k$ activation weights.
 
 \subsubsection{Uncertainty Maps}
 Although the previous approach pulls regular saliency maps towards uncertainty-informed saliency maps, they still do not represent pure uncertainty distribution on the input images. Thus, we formulate uncertainty maps by calculating predictive uncertainty over MC samples, and then map the gradient information from the predictive uncertainty back to input pixels.  We define per-pixel uncertainty-based saliency in Eq.~\ref{eq:unmap}, following our notation in Eq.~\ref{eq:PU}.
\begin{equation}
\begin{aligned}\label{eq:unmap}
    s_{ij}&= \frac{\partial H(y|x, D)}{\partial x_{ij}}
    \end{aligned}
\end{equation}
This gradient-based approach reveals spatial patterns of detector confidence, identifying regions where the model exhibits high uncertainty about its predictions.




\section{Analysis}
\label{sec:exp}

We conduct experiments on uncertainty of deepfake (source) detectors, region- and pixel-based uncertainty, uncertainty estimation techniques, with ablation studies.

\subsection{Uncertainty of Deepfake Detectors}

\begin{table}[!ht]
    \begin{small}
	\begin{center}
		
		\begin{tabular}{ll|c|c|c|c|c|c}
		    \hline
		    \multicolumn{1}{l}{$M$} & \multicolumn{1}{l}{\hspace{-0.1in}Eval} &
            \multicolumn{1}{c}{DF} & \multicolumn{1}{c}{F2F} & \multicolumn{1}{c}{FSh} & \multicolumn{1}{c}{FSw} & \multicolumn{1}{c}{NT} & \multicolumn{1}{c}{All}\\
		    \hline \hline
		      R & \hspace{-0.1in} acc & 96.96 & 93.96 & 99.15 & 92.35 & 94.51 & 94.08 \\
		    \hline
		     R$_B$& \hspace{-0.1in} acc & 96.38 & 94.91 & 98.81 & 93.64 & 93.39 & 95.32\\
                  \cline{2-8}
		       & \hspace{-0.1in} PU & 0.075 & 0.077 & 0.043 & 0.097 & 0.069 & 0.037\\
                  & \hspace{-0.1in} MU & 0.031 & 0.028 & 0.026 & 0.051 & 0.038 & 0.018\\
		    \hline
      	E & \hspace{-0.1in} acc & 99.96 & 99.28 & 99.08 & 99.42 & 99.67 & 99.38 \\
		    \hline
		     E$_B$ & \hspace{-0.1in} acc & 95.87 & 93.24 & 98.82 & 97.75 & 92.31 & 90.93 \\
                  \cline{2-8}
		       & \hspace{-0.1in} PU & 0.209 & 0.151 & 0.203 & 0.168 & 0.246 & 0.263\\
                  & \hspace{-0.1in} MU & 0.098 & 0.092 & 0.107 & 0.095 & 0.132 & 0.128\\
		    \hline
      	F & \hspace{-0.1in} acc & 96.73 & 95.12 & 95.65 & 96.04 & 93.31 & 96.14 \\
		    \hline
		     F$_B$& \hspace{-0.1in} acc & 96.30 & 94.37 & 95.52 & 95.76 & 91.59 & 95.77 \\
                  \cline{2-8}
		       & \hspace{-0.1in} PU & 0.015 & 0.026 & 0.056 & 0.016 & 0.089 & 0.028 \\
                  & \hspace{-0.1in} MU & 0.001 & 0.003 & 0.008 & 0.002 & 0.006 & 0.002 \\
		    \hline
      	M & \hspace{-0.1in} acc & 97.54 & 93.50 & 97.63 & 97.71 & 87.83 & 88.19 \\
		    \hline
		    M$_B$& \hspace{-0.1in} acc & 94.16 & 84.91 & 95.91 & 92.95 & 77.71 & 87.40 \\
                  \cline{2-8}
		       & \hspace{-0.1in} PU & 0.083 & 0.147 & 0.071& 0.103 & 0.241 & 0.182 \\
                  & \hspace{-0.1in} MU & 0.007 & 0.007 & 0.008 & 0.007 & 0.006 & 0.007 \\
		    \hline
		\end{tabular}
		
	\end{center}
    \caption{Accuracy and predictive (PU) and model (MU) uncertainties for (R)esnet18~\cite{resnet}, (E)fficientNet~\cite{efficientnet}, (F)akeCatcher~\cite{FakeCatcher}, and (M)otion-based detector~\cite{motion}, with their Bayesian $X_B$  variants; per generator in FaceForensics++~\cite{FF++}.}\label{tab:tab2}
	\end{small}
\end{table}

Tab.\ref{tab:tab2} presents binary classification performance for each synthetic content generator in the FF dataset. For instance, the DF column displays results from models trained and evaluated exclusively on DF-generated samples versus authentic content using the aforementioned split. The rightmost column combines all synthetic categories against real samples in a unified binary classification task.

\begin{table}[!h]
	\begin{center}
		
		\begin{tabular}{l l|c}
		    \hline
		    \multicolumn{1}{l}{$M$odels} & \multicolumn{1}{l}{Eval} &
             \multicolumn{1}{c}{Results} \\
		    \hline \hline
		    Resnet18 & accuracy  & 94.23 \\
		    \cline{1-3}
		    Resnet18$_B$& accuracy  & 93.54 \\
                  \cline{2-3}
		      & predictive uncertainty & 0.119 \\
               & model uncertainty & 0.054 \\
		    \hline
      	 FakeCatcher & accuracy & 97.99 \\
		    \cline{1-3}
		 FakeCatcher$_B$ & accuracy & 98.21 \\
                  \cline{2-3}
		     & predictive uncertainty & 0.030\\
                 & model uncertainty & 0.009\\
                  \hline
		\end{tabular}
        \caption{Accuracy and uncertainty results of regular/Bayesian variants of blind/biological detectors on FAVC.}
		\label{tab:FACVresults}
	\end{center}
\end{table}

\begin{figure*}[ht!]
\begin{center}
\includegraphics[width=0.76\linewidth]{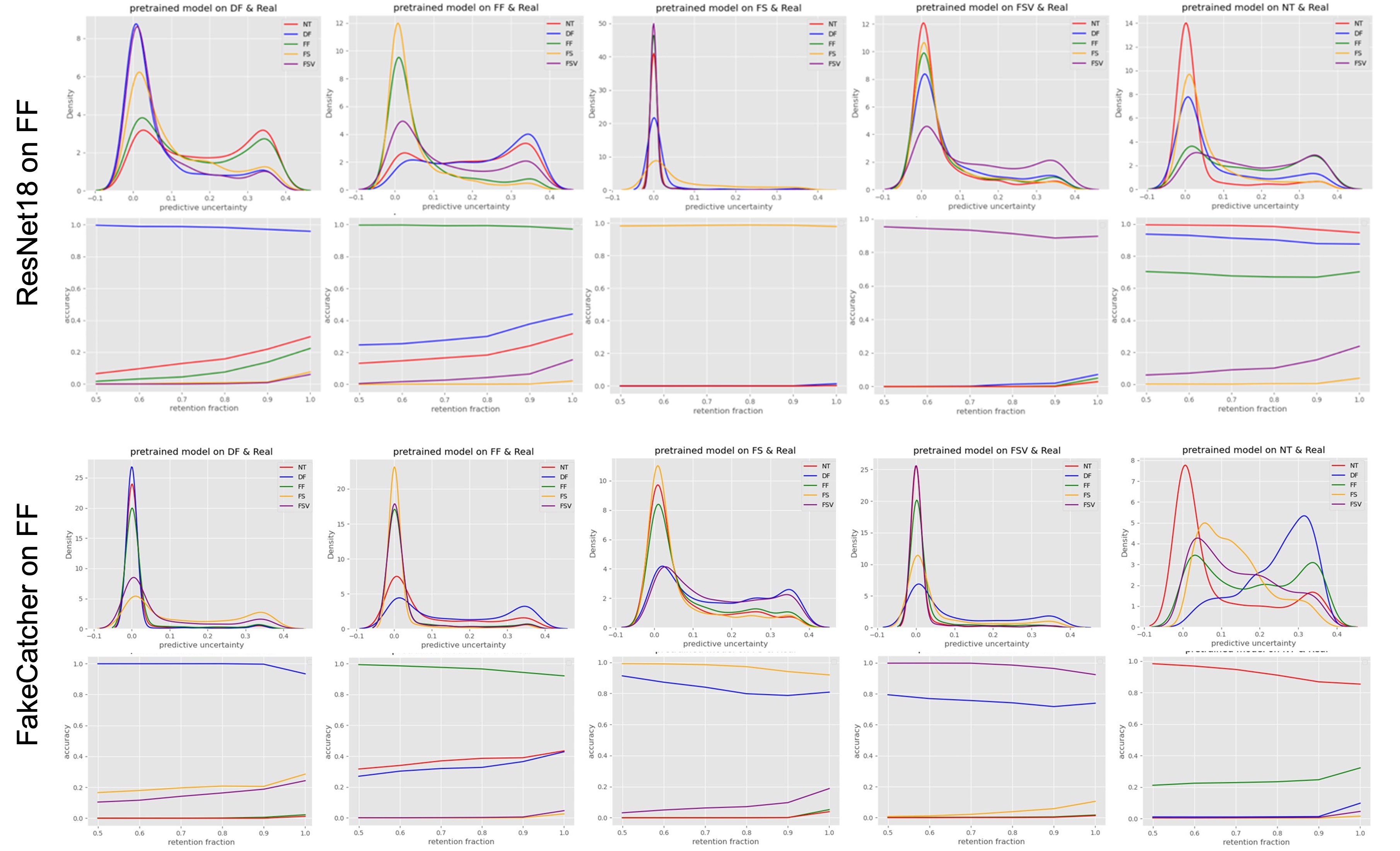}
\end{center}
\caption{Each column shows density histograms and accuracy retention curves for 5 generators in FF, trained and tested per generator.} \label{fig:fig2}
\end{figure*}
The findings reveal a notable performance gap between traditional neural networks and their Bayesian counterparts: while complex ensemble architectures achieve high accuracy rates (99.38\% in row 4), their BNN implementations experience substantial accuracy degradation, exceeding 8\% (90.93\% in row 5). In contrast, biological detectors maintain remarkable stability, with accuracy drops below 1\% (comparing rows 7-8 and 10-11). This performance differential is further substantiated by the uncertainty metrics, where blind models exhibit considerably higher predictive and model uncertainties (0.263 PU and 0.128 MU versus 0.028 PU and 0.002 MU in rows 6 and 9, respectively). Moreover, more complex blind detectors show greater accuracy degradation and higher uncertainty with Bayesian variants. In Tab.~\ref{tab:FACVresults}, we repeat the evaluation on FAVC dataset with 3 generators and obtain similar insights, especially for validating the certainty of biological detectors.

 Fig.~\ref{fig:fig2} shows density histograms and corresponding retention curves, which are computed by testing ResNet18 and FakeCatcher for real/fake detection with five generators in FF. 
 We observe that (1) biological detectors have a narrower variance of uncertainty in this binary setting, (2) similar face manipulations provide relatively better generalizability (closer curves for DF, FSh, and FSw vs. NT and F2F) and (3) for biological detectors, per-generator models can generalize to similar fakes (last row, cols. 3-4). For both detector types, we observe a unique behavior for NT, suggesting fundamental challenges in this synthesis paradigm.

\begin{table}[!ht]
    \begin{small}
	\begin{center}
		
		\begin{tabular}{l l|c|c|c|c|c}
		    \hline
            \multicolumn{1}{l}{$M$} & \multicolumn{1}{l}{Eval} &
            \multicolumn{1}{c}{L$_{DF}$} & \multicolumn{1}{c}{L$_{F2F}$} & \multicolumn{1}{c}{L$_{FSh}$} & \multicolumn{1}{c}{L$_{FSw}$} & \multicolumn{1}{c}{L$_{NT}$} \\
		    \hline \hline
		      R$_B$& acc & 97.75 & 91.25 & 46.75 & 18.25 & 70.92 \\
                  \cline{2-7}
		       & PU & 0.074 & 0.179 & 0.246 & 0.193 & 0.252 \\
              & MU & 0.036 & 0.139 & 0.142 & 0.090 & 0.151 \\
		    \hline
		    F$_B$& acc & 96.14 & 70.27 & 71.91 & 83.43 & 67.86 \\
                  \cline{2-7}
		     & PU & 0.143 & 0.219 & 0.231 & 0.225 & 0.215 \\
                & MU & 0.003 & 0.015 & 0.008 & 0.007 & 0.013 \\
		    \hline
            M$_B$& acc & 93.91 & 64.50 & 82.17 & 52.00 & 69.75 \\
                  \cline{2-7}
		     & PU & 0.236 & 0.271 & 0.271 & 0.256 & 0.253 \\
                & MU & 0.008 & 0.009 & 0.011 & 0.009 & 0.010 \\
		    \hline
		\end{tabular}
        \caption{Accuracy and predictive (PU) and model (MU) uncertainties for Bayesian (R)esnet18~\cite{resnet}, (F)akeCatcher~\cite{FakeCatcher}, and (M)otion-based detector~\cite{motion}, with leave-one-out trainings in FF.}
		\label{tab:LOO}
	\end{center}
	\end{small}
\end{table}
\begin{figure}[hb!]
\begin{center}
\includegraphics[width=1\linewidth] {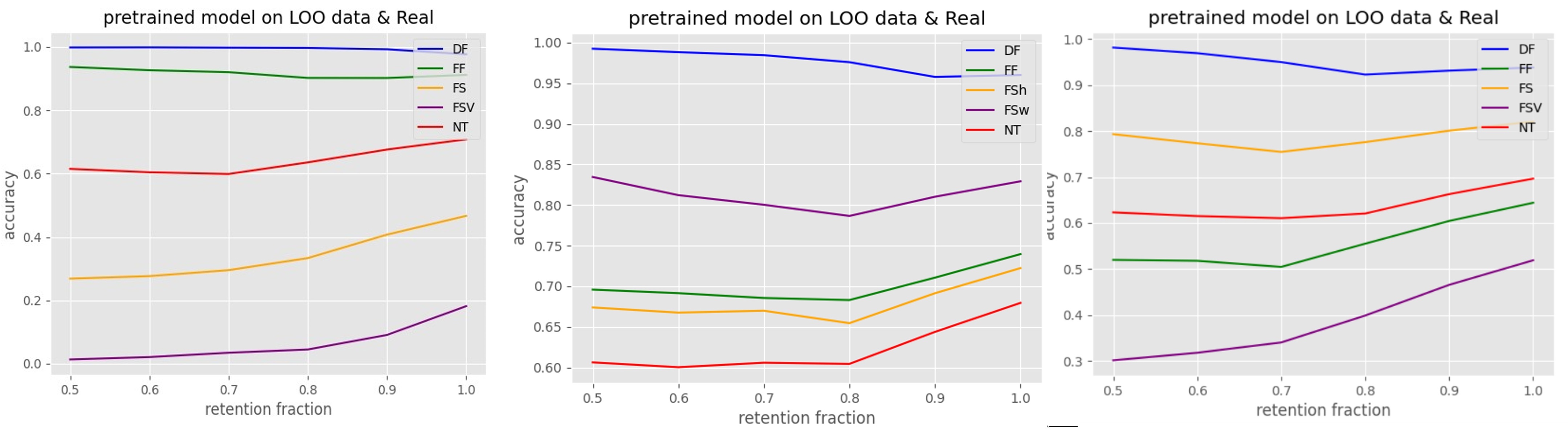}
\end{center}
\caption{Retention plots of (a) ResNet18, (b) FakeCatcher, (c) motion-based detector on FF for DF, following Tab.~\ref{tab:LOO}.} \label{fig:LOO}
\end{figure}

To mimic a production environment, we conduct leave-one-out (LOO) experiments (Tab. \ref{tab:LOO}). For LOO, we use the same split for train/validation on five classes and test the best model's accuracy on the left-out class's test set. Overall, generalizing to FSw's artifacts is harder, however FakeCatcher can achieve it. Another interpretation of Tab.~\ref{tab:LOO} is that generalization capability increases as detector uses more modalities, from spatial (blind) to spatio-temporal (motion) to spectro-temporal (PPG) representations.

Fig.~\ref{fig:LOO} visualizes retention curves of the LOO experiment in the first column. Retention fraction represents percentage of retained data based on predictive uncertainty, with the expectation of uncertain samples decreasing accuracy for calibrated models. Here, $LOO_{DF}$ (blue curve) refers to the use of F2F, FSh, FSw, NT, and real data in training and validation, and DF data only for testing. Each column shows retention curves for the corresponding model in Tab.~\ref{tab:LOO}. Retention curves confirm our findings about generalization.

\subsection{Uncertainty of Deepfake Source Detection}
Tab.~\ref{tab:fakesource} shows uncertainty and accuracy on FF for deepfake source detection. Source detection observations are similar: Complex and large networks overfit and their Bayesian versions cannot reproduce the same accuracy, exhibit dramatic accuracy drops for source detection (EfficientNet: 99.46\% $\rightarrow$ 89.77\%) with extremely high uncertainty (middle blocks). Smaller networks (ResNet18, first block) and biological detectors (FakeCatcher, last block)  maintain relatively stable performance and low uncertainty.

\begin{table}[!ht]
    \begin{small}
	\begin{center}
		
		\begin{tabular}{ll|c|c|c|c|c|c}
		    \hline
		   \multicolumn{1}{l}{$M$} & \multicolumn{1}{l}{\hspace{-0.1in}Eval} &
            \multicolumn{1}{c}{DF} & \multicolumn{1}{c}{F2F} & \multicolumn{1}{c}{FSh} & \multicolumn{1}{c}{FSw} & \multicolumn{1}{c}{NT}& \multicolumn{1}{c}{All}\\
		    \hline \hline
		      R & \hspace{-0.1in} acc & 98.58
 & 96.66 & 98.16 & 95.5 & 92.66 & 96.32\\
		    \cline{1-8}
		   R$_B$& \hspace{-0.1in} acc & 97.83 & 97.41
 & 97.33 & 96.75 & 93.66 & 95.95 \\
                  \cline{2-8}
		       & \hspace{-0.1in} PU & 0.055 & 0.068 & 0.134 & 0.103 & 0.108 & 0.131\\
                  & \hspace{-0.1in} MU & 0.024 & 0.033 & 0.082 & 0.060 & 0.055 & 0.074\\
		    \hline
      	 X & \hspace{-0.1in} acc & 99.83 & 99.41 & 98.92 & 99.00 & 99.00 & 99.17 \\
		    \cline{1-8}
		  X$_B$& \hspace{-0.1in} acc & 97.08 & 98.25 & 91.41 & 95.58 & 99.91 & 89.37\\
                  \cline{2-8}
		      & \hspace{-0.1in} PU & 0.257 & 0.109 & 0.518 & 0.388 & 0.054 & 0.344\\
                 & \hspace{-0.1in} MU & 0.160 & 0.075 & 0.366 & 0.266 & 0.031 & 0.227\\
		    \hline
              E & \hspace{-0.1in} acc & 99.91 & 99.08 & 98.92 & 99.75 & 99.33 & 99.46\\
		    \cline{1-8}
		 E$_B$& \hspace{-0.1in} acc & 82.75 & 88.75 & 85.75 & 94.33 & 92.5 & 89.77\\
                  \cline{2-8}
		  & \hspace{-0.1in} PU & 0.984 & 0.806 & 1.091 & 0.714 & 0.844 & 0.894\\
                  & \hspace{-0.1in} MU & 0.437 & 0.444 & 0.571 & 0.382 & 0.372 & 0.432\\
		    \hline
             B & \hspace{-0.1in} acc & 99.91 & 99.33 & 99.00 & 99.58 & 99.08 & 99.38 \\
		    \cline{1-8}
		  B$_B$ & \hspace{-0.1in} acc & 87.41 & 91.83 & 93.58 & 97.91 & 84.00 & 91.08\\
                  \cline{2-8}
		     & \hspace{-0.1in} PU & 1.328 & 1.021 & 1.210 & 0.824 & 1.259 & 1.154\\
                 & \hspace{-0.1in} MU & 0.326 & 0.449 & 0.535 & 0.402 & 0.447 & 0.447\\
		    \hline
              F & \hspace{-0.1in} acc & 92.08 & 90.31 & 92.41 & 90.97 & 85.27 & 91.26\\
		    \cline{1-8}
		  F$_B$& \hspace{-0.1in} acc & 93.58 & 88.34 & 93.20 & 91.67 & 80.98 & 90.18\\
                  \cline{2-8}
		   & \hspace{-0.1in} PU & 0.125 & 0.223 & 0.122 & 0.139 & 0.310 & 0.198\\
                  & \hspace{-0.1in} MU & 0.004 & 0.015 & 0.006 & 0.005 & 0.032 & 0.013 \\
		    \hline
		\end{tabular}
		\caption{Accuracy and uncertainty results of regular and Bayesian detectors (R)esnet18~\cite{resnet}, (X)ception~\cite{xception},  (E)fficientNet~\cite{efficientnet},
        Mo(B)ileNet~\cite{mobilenetv2}, (F)akeCatcher~\cite{FakeCatcher} for source detection on real and five generators in FF.}
        \label{tab:fakesource}
	\end{center}
	\end{small}
\end{table}

\subsection{Region-based Uncertainty Analysis}
In order to couple generator types per face manipulations to detector uncertainty, we conduct region-based experiments for deepfake detection (left half) and source detection (right half) in Tab.~\ref{tab:multiclass}. 
For example, when the training samples exclude bottom half of the faces, source detection for NT shows dramatic performance drop (95.81\% $\rightarrow$ 91.68\%), confirming mouth-centric manipulation focus. Similarly for F2F, removing symmetry elements from the training set (half mouth or one eye) reduces its source detection, as F2F is a mask-based technique creating symmetric priors. Region-based results also indicate that uncertainty measures are highly correlated with accuracy measures. It is also observed that each generator exhibits characteristic regional uncertainty patterns that can serve as forensic signatures. 

\begin{table*}[!h]
    \begin{small}
	\begin{center}

		\begin{tabular}{lll|c|c|c|c|c|c||c|c|c|c|c|c}
		    \hline
		    \multicolumn{1}{l}{Region} & \multicolumn{1}{l}{\hspace{-0.05in}$M$} & \multicolumn{1}{l}{\hspace{-0.15in}Eval} &
            \multicolumn{1}{c}{DF} & \multicolumn{1}{c}{F2F} & \multicolumn{1}{c}{FSh} & \multicolumn{1}{c}{FSw} & \multicolumn{1}{c}{NT}& \multicolumn{1}{c}{All} &
            \multicolumn{1}{c}{DF} & \multicolumn{1}{c}{F2F} & \multicolumn{1}{c}{FSh} & \multicolumn{1}{c}{FSw} & \multicolumn{1}{c}{NT}& \multicolumn{1}{c}{All}\\
		    \hline \hline
               & R & \hspace{-0.1in}acc& 99.45 & 98.20 & 99.30 & 99.10 & 97.76 & 97.86 & 96.25 & 95.25 & 97.00 & 90.83 & 88.83 & 93.65\\
		    \cline{2-15}
                \multirow{2}{*}{\includegraphics[width=0.12\linewidth]{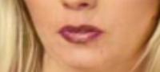}\hspace{-0.1in}} & R$_B$& \hspace{-0.1in}acc& 98.37 & 97.69 & 99.17 & 97.03 & 92.17 & 95.00 & 95.25 & 93.33
                & 98.00 & 91.33 & 92.75 & 93. 25\\
                  \cline{3-15}
		      & & \hspace{-0.1in}PU& 0.133 & 0.077 & 0.047 & 0.103 & 0.124 & 0.058 & 0.096 & 0.132 & 0.054 & 0.187 & 0.144 & 0.154 \\
                &  & \hspace{-0.1in}MU& 0.056 & 0.045 & 0.026 & 0.058 & 0.079 & 0.035 & 0.050 & 0.062 & 0.029 & 0.092 & 0.067 & 0.076 \\
                  \hline
                 & R & \hspace{-0.1in}acc& 99.30 & 96.46 & 97.86 & 95.02 & 95.77 & 97.56 & 93.33 & 92.08 & 96.16
                & 90.83 & 88.58 & 89.64 \\
		    \cline{2-15}
                \multirow{2}{*}{\includegraphics[width=0.06\linewidth]{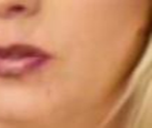}\hspace{-0.1in}} & R$_B$& \hspace{-0.1in}acc& 97.63 & 95.27 & 97.99 & 94.37 & 88.03 & 94.45& 94.75 & 89.58
                & 95.91 & 86.16 & 87.41 & 88.99 \\
                  \cline{3-15}
		      & & \hspace{-0.1in}PU& 0.085 & 0.099 & 0.053 & 0.101 & 0.145 & 0.080 & 0.140 & 0.192 & 0.112 & 0.267 & 0.192 & 0.217 \\
                & & \hspace{-0.1in}MU& 0.051 & 0.053 & 0.034 & 0.069 & 0.095 & 0.049 & 0.077 & 0.108 & 0.061 & 0.152 & 0.105 & 0.121 \\
                  \hline
                & R &\hspace{-0.1in}acc& 99.54 & 97.27 & 98.84 & 98.05 & 95.81 & 97.32 & 95.42 & 91.5 & 97.58
                & 93.33 & 88.16 & 91.88 \\
		    \cline{2-15}
                \multirow{2}{*}{\includegraphics[width=0.12\linewidth]{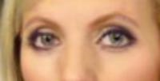}\hspace{-0.1in}} & R$_B$&\hspace{-0.1in}acc& 96.04 & 96.65
                & 98.21 & 94.67 & 84.47 & 92.48 & 94.33 & 90.75
                & 97.75 & 91.16 & 87.58 & 91.37 \\
                  \cline{3-15}
		      & & \hspace{-0.1in}PU& 0.101 & 0.111 & 0.087 & 0.115 & 0.154 & 0.089 & 0.114 & 0.143 & 0.060 & 0.170 & 0.193 & 0.162 \\
                & &\hspace{-0.1in}MU& 0.063 & 0.072 & 0.063 & 0.069 & 0.093 & 0.063 & 0.057 & 0.073 & 0.031 & 0.085 & 0.103 & 0.083 \\
                  \hline
                & R &\hspace{-0.1in}acc& 98.95 & 95.53 & 98.86 & 95.88 & 91.68 & 95.08 & 93.58
 & 80.41 & 96.00 & 84.16 & 73.66 & 83.56 \\
		    \cline{2-15}
                \multirow{2}{*}{\includegraphics[width=0.06\linewidth]{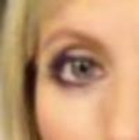}\hspace{-0.1in}} & R$_B$& \hspace{-0.1in}acc& 96.64 & 91.68 & 98.19 &
                92.58 & 79.80 & 83.78 & 88.91 & 78.83
                & 93.42 & 84.16 & 78.5 & 83.39\\
                  \cline{3-15}
		      & &\hspace{-0.1in}PU& 0.093 & 0.127 & 0.044 & 0.106 & 0.140 & 0.250 & 0.168 & 0.277 & 0.167 & 0.254 & 0.266
 & 0.247 \\
                & & \hspace{-0.1in}MU& 0.055 & 0.076 & 0.030 & 0.067 & 0.088 & 0.054 & 0.082 & 0.138 & 0.091 & 0.130 & 0.131 & 0.125\\
                  \hline
  \end{tabular}
				\caption{Region-based analysis of accuracy and predictive (PU) and model uncertainty (MU) for deepfake detection (left half) and source detection (right half) on FF, using Resnet18 (R) and BNN Resnet18 (R$_B$).}\label{tab:multiclass}
	\end{center}
	\end{small}
\end{table*}

\section{Uncertainty Maps for Deepfake Detection}
\begin{figure}[h!]
\begin{center}
\includegraphics[width=1\linewidth] {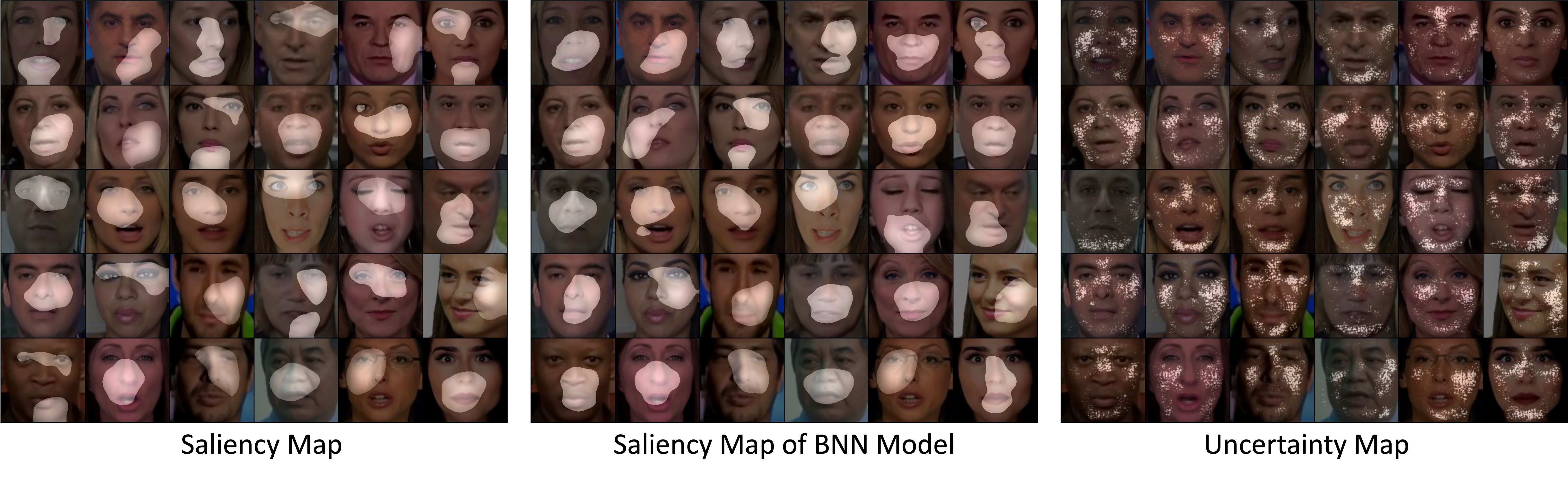}
\end{center}
\caption{Saliency, Bayesian saliency, and uncertainty maps of ResNet18 detector on NT samples.} \label{fig:pixel}
\end{figure}

We visually compare saliency map of ResNet18 detector, its Bayesian saliency, and its uncertainty map in Fig.~\ref{fig:pixel}, for NT.

Our uncertainty maps reveal interpretable spatial patterns that complement traditional saliency analysis while providing novel insights. 
\textit{From a generator perspective}, uncertainty concentrates around mouth regions and facial boundaries for NT, creating distinctive "skull-like" patterns in cheek and forehead areas where fewer artifacts exist. For DF, uncertainty localizes primarily around facial perimeters and identity-critical regions, reflecting the encoder-decoder compression artifacts. F2F exhibits symmetric uncertainty patterns consistent with mask-based manipulation approaches. \textit{From an interpretability angle,} uncertainty maps effectively identify regions where detection confidence decreases, often corresponding to subtle artifact boundaries. High uncertainty regions frequently correspond to areas where biological constraints are violated. While saliency maps highlight discriminative features, uncertainty maps reveal areas requiring additional evidence for confident classification. Lastly \textit{for a comparative view}, traditional saliency focuses on obvious discriminative artifacts. Bayesian saliency produces more diffuse, averaged activations that create blob-like patterns in central facial regions. This is also expected considering that noise, mouth, and middle areas contain most of the artifacts. Uncertainty maps generate distinctive spatial patterns highlighting regions of low detector confidence, particularly in peripheral facial areas. This multi-modal visualization provides comprehensive insights into both what detectors focus on (saliency) and where they lack confidence (uncertainty), enabling more informed deployment decisions.

\section{Ablation Studies}
\label{ablation}
Tab. \ref{tab:Test10} shows the impact of parameter values on BNN performance. Exp.1, Exp.2, Exp.3, Exp.4 and Exp.5 refer to the parameter settings of $n=\{40, 10, 40, 40, 40\},  \delta_{moped}=\{0.1, 0.1, 0.5, 0.1, 0.1\}, kl_{factor}=\{1,1,1,0.5,0.1\}$
, respectively for number of MC samples $n$, moped delta value $\delta_{moped}$, and scaling coefficient for KL loss $kl_{factor}$.
The results show that increasing $n$ from 10 to 40  marginally improves uncertainty quality without substantial accuracy changes. Smaller $kl_{factor}$ causes degradation for NT and the quality of uncertainty measures in general deepfake detection. Finally, increasing $\delta_{moped}$ causes a significant drop for BNN performance so it should be fine-tuned.

\begin{table}[!ht]
    \begin{small}
	\begin{center}
		
		\begin{tabular}{l l|c|c|c|c|c}
		    \hline
		    \multicolumn{1}{l}{Dataset} & \multicolumn{1}{l}{Eval} &
            \multicolumn{1}{c}{Exp.1} & \multicolumn{1}{c}{Exp.2} & \multicolumn{1}{c}{Exp.3} & \multicolumn{1}{c}{Exp.4} &
            \multicolumn{1}{c}{Exp.5} \\
		    \hline \hline
               \multirow{3}{*}{All} & acc & 96.72 & 96.94 & 87.79 & 96.60 & 97.09\\
		    \cline{2-7}
                & PU & 0.042 & 0.052 & 0.227 & 0.058 & 0.050\\
		      & MU & 0.025 & 0.026 & 0.120 & 0.036 & 0.029\\
                  \hline
                \multirow{1}{*}{Neural} & acc & 93.61 & 93.79 & 84.50 & 93.91 & 90.92\\
		    \cline{2-7}
                \multirow{1}{*}{Textures} & PU & 0.127 & 0.119 & 0.301 & 0.114 & 0.129\\
		      & MU & 0.072 & 0.063 & 0.175 & 0.075 & 0.080\\
            \hline
  \end{tabular}\caption{Impact of the hyperparameters on BNN performance.} 		
  \label{tab:Test10}
  	\end{center}
    \end{small}
\end{table}

Tab. \ref{tab:Test11} shows that increasing $dr$ (dropout ratio) causes significant drop in MC dropout performance as measured on NN and F2F. On the other hand, a fine-tuned dropout ratio may even result in an improved accuracy.

\begin{table}[!ht]
    \begin{small}
	\begin{center}
 
		\begin{tabular}{l|l|c|c|c|c|c}
		    \hline
		   \multicolumn{1}{l}{Data} & \multicolumn{3}{c}{NT} &
            \multicolumn{3}{c}{F2F} \\\hline \hline
          Eval & \multicolumn{1}{l}{dr=0.2} & \multicolumn{1}{c}{0.3}  & 0.5 & \multicolumn{1}{c}{dr=0.2} & \multicolumn{1}{c}{0.3}  & 0.5 \\
		    \hline
          acc & 97.75 & 97.17 & 50.27 & 98.54 & 98.99 & 49.84\\
              MU & 0.015 & 0.026 & 0.030 & 0.013 & 0.011 & 0.023\\
            \hline
  \end{tabular} \caption{$dr$ impact on Resnet18 performance with NT and F2F.} 
		\label{tab:Test11}
	\end{center}
    \end{small}
\end{table}

Lastly, we measure the robustness of Bayesian detectors on adversarial samples. Tab.~\ref{tab:adv} reports accuracy of the attacked BNN ResNet18 before and after adversarial generation on five generator subsets, reducing the detection accuracy by 93.53\% on the average.

\begin{table}[!ht]

	\begin{center}
		
\begin{tabular}{llllll}
\hline
\multicolumn{1}{l|}{Generator}            & \multicolumn{1}{l|}{DF}      & \multicolumn{1}{l|}{F2F}      & \multicolumn{1}{l|}{FSw}     & \multicolumn{1}{l|}{NT}      & FSh      \\ \hline \hline
\multicolumn{1}{l|}{Baseline}        & \multicolumn{1}{l|}{95.81} & \multicolumn{1}{l|}{95.69} & \multicolumn{1}{l|}{89.58} & \multicolumn{1}{l|}{93.64} & 97.26 \\ \hline
\multicolumn{1}{l|}{After attack} & \multicolumn{1}{l|}{0.30}  & \multicolumn{1}{l|}{0.03}  & \multicolumn{1}{l|}{3.07}  & \multicolumn{1}{l|}{0.94}  & 0     \\
                            
\hline
\end{tabular}\caption{Adversarial robustness of BNN Resnet18 detector.}
		\label{tab:adv}
	\end{center}
\end{table}

\section{Implementation Details}\label{app:imp}

\begin{figure}[ht!]
\begin{center}
\includegraphics[width=0.9\linewidth] {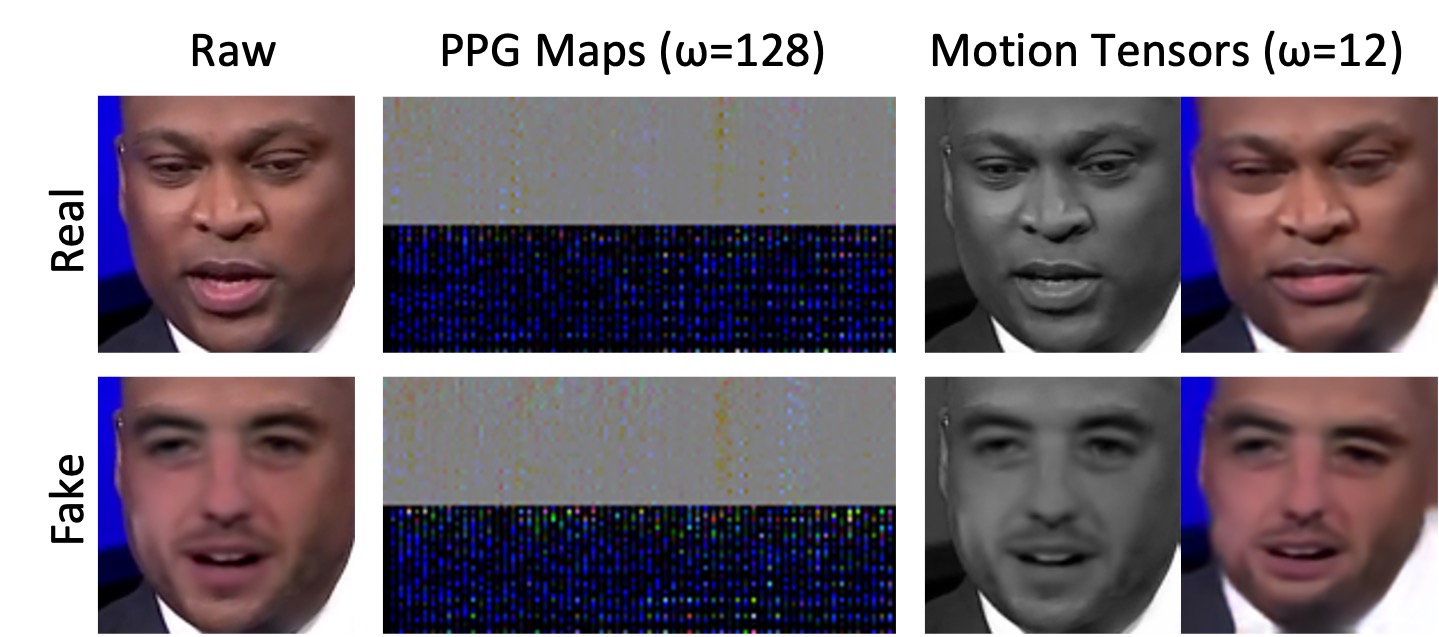}
\end{center}
\caption{Data representations created by different deepfake detectors for a pair of real and fake videos from DF in FF.} \label{appf:reps}
\end{figure}

The first four detectors consume raw data whereas the last two detectors exploit intermediate representations. FakeCatcher~\citep{FakeCatcher} extracts photoplethysmography (PPG) maps from videos, representative of spatial, temporal, and spectral signal behavior of heart rates. We follow their construction of PPG maps, except setting $\omega=64$. Motion-based detector~\citep{motion} extracts dual representations to represent submuscular motion by deep and phase-based motion magnification. We follow their construction of motion tensors with the optimum suggested parameters. Intermediate representations for three types of detectors (raw, PPG-based, and motion-based) are sampled in Fig.~\ref{appf:reps}. For network counterparts, we use VGG19~\citep{vgg} and C3D~\citep{c3d} respectively, as suggested in~\cite{ijcb} and~\cite{motion}. 


For our BNN implementations using Bayesian repo~\citep{krishnan2022bayesiantorch}, prior parameters are set as $\mu_{prior}=0$, $\sigma_{prior}=1$, $\mu i_{posterior}=0$, and $\rho i_{posterior}=-3$. \textit{moped} is enabled with $\delta_{moped}=0.1$ selecting \textit{reparameterization} type.

During training, models are trained with Adam optimizer~\citep{kingma2014adam} with a learning rate (LR) of 0.0001 for all architectures, except C3D. C3D LR is initiated as 0.001 and dynamic LR is applied with 0.1 scaling after each 10 epoch of overall 100 epochs. All other models are trained for 200 epochs. All weights are initiated using pretrained models on Imagenet \citep{imagenet} from torchvision~\citep{torchvision}, except C3D model which is pretrained on UCF101 dataset~\citep{soomro2012ucf101}. 

The lowest validation loss model is selected as the best model. Since accuracy is computed using $n$ MC samples, our definition of the best model may not always correspond to the model with the highest accuracy. MC sampling enables variations at the output that may cause some noise in accuracy. Predictive and model uncertainties represent the average uncertainty measures of the test splits.

For saliency construction, batch size is set as 1 and the cut-offs are set as 20\%, 20\%, and 10\% experimentally for saliency, Bayesian saliency, and uncertainty maps. 


\section{Limitations and Discussion}
Our comprehensive analysis establishes several critical insights that fundamentally challenge current deepfake detection evaluation paradigms. (1) Biological detectors demonstrate markedly superior uncertainty calibration compared to blind approaches, maintaining stable performance during Bayesian conversion while exhibiting substantially lower uncertainty levels. This suggests that incorporating domain-specific physiological priors enhances not only detection accuracy but also prediction reliability, a crucial consideration for deployment. (2) The strong correlation between uncertainty measures and generalization performance across unseen generators establishes uncertainty quantification as a fundamental requirement rather than auxiliary information. Systems exhibiting high uncertainty should trigger additional verification procedures, preventing overconfident misclassifications in critical applications.
(3) Uncertainty patterns encode generator-specific signatures that enable forensic analysis beyond binary classification. These findings suggest that uncertainty-aware systems could provide valuable attribution capabilities for investigating deepfake origins and understanding manipulation techniques employed.
(4) Complex architectures without appropriate inductive biases exhibit poor uncertainty calibration, highlighting the importance of domain-informed design. The dramatic performance degradation suggests that model complexity alone does not guarantee reliable uncertainty estimation. (5) Our adversarial evaluation reveals concerning vulnerabilities across all detector types, with performance drops exceeding 99\% under simple gradient-based attacks. 
The relationship between baseline uncertainty and adversarial susceptibility could open new pathways for uncertainty-aware systems to implement adaptive security measures, increasing verification for elevated uncertainty levels.

Unfortunately, Bayesian inference introduces substantial computational costs through multiple forward passes and weight sampling. Future work should explore more efficient uncertainty estimation techniques, potentially through distillation or approximation methods. While our analysis spans multiple generators, evaluation on completely novel synthesis paradigms (e.g., diffusion-based deepfakes) remains an open question requiring continued investigation.

\section{Conclusion}
We propose an in-depth analysis of deepfake detectors, generators, and source-detectors from an uncertainty perspective; including region-based detection experiments, novel uncertainty maps, blind and biological detector comparisons, and revelations between detector architectures and generator artifacts. Uncertainty analysis in the deepfake landscape is a new but essential dimension before releasing these detectors for public use. We have demonstrated that underconfident certain models are superior to overconfident uncertain models in terms of generalization. Our results indicate that generator artifacts can guide both detection and source detection, in image, region, and pixel levels.

As future work, we would like to build source detectors incorporating uncertainty maps, as they are shown to be information-rich for this task. As generative models and their applications become more ubiquitous and embedded in our lives, we support the responsible dissemination of tools that foster explainability, transparency, trust, and risk-awareness to ensure their use for future social good.



{
    \small
    \bibliographystyle{ieeenat_fullname}

}

\end{document}